\documentclass[letterpaper]{article} 
\usepackage{aaai25}  
\usepackage{times}  
\usepackage{helvet}  
\usepackage{courier}  
\usepackage[hyphens]{url}  
\usepackage{graphicx} 
\usepackage{natbib}  
\usepackage{caption} 
\usepackage{amsfonts}
\usepackage{amsmath}
\usepackage[ruled]{algorithm2e}
\SetKwComment{Comment}{// }{}
\usepackage{booktabs}

\title{Extracting Interpretable Task-Specific Circuits\\from Large Language Models for Faster Inference}
\author{
    Jorge García-Carrasco,
    Alejandro Maté,
    Juan Trujillo
}
\affiliations{
    Department of Software and Computing Systems\\ 
    University of Alicante, Spain\\

    jorge.g@ua.es, \{amate,jtrujillo\}@dlsi.ua.es
}

\usepackage{bibentry}

\begin{document}

\maketitle

\begin{abstract}
Large Language Models (LLMs) have shown impressive performance across a wide range of tasks. However, the size of LLMs is steadily increasing, hindering their application on computationally constrained environments. On the other hand, despite their general capabilities, there are many situations where only one specific task is performed, rendering all other capabilities unnecessary and wasteful. This leads us to the following question: \emph{Is it possible to extract the minimal subset from an LLM that is able to perform a specific task in a faster, standalone manner?} Recent works on Mechanistic Interpretability (MI) have shown that specific tasks are performed by a localized subset of components, or circuit. However, current techniques used to identify the circuit cannot be used to extract it for its standalone usage. In this work, we propose a novel approach to automatically extract the subset of the LLM that properly performs a targeted task requiring no additional training and a small amount of data samples. We evaluate our approach on different tasks and show that the resulting models are (i) considerably smaller, reducing the number of parameters up to $82.77\%$ and (ii) more interpretable, as they focus on the circuit that is used to carry out the specific task, and can therefore be understood using MI techniques.
\end{abstract}

%

\section{Introduction}

Large Language Models (LLMs) based on the Transformer architecture \citep{vaswani2017attention} have recently shown impressive performance across a wide range of tasks \citep{brown2020language,achiam2023gpt}. These models are characterized by having a large amount of parameters as well as being trained on large amounts of broad data in a self-supervised manner (e.g predicting the next token). In fact, it has been empirically shown that the performance of LLMs increases with the number of parameters, as well as training dataset size and amount of training \citep{kaplan2020scaling}. Therefore, we expect the next generation of LLMs to improve their performance even more as they continue growing even larger.

However, the computational cost can become prohibitive in many situations. As an example, it takes around 130.4GB of memory just to load the largest Llama2 model \citep{touvron2023llama} (70 billion parameters in 16-bit precision), totaling up to two A100 80GB GPUs. This clearly shows that the use of LLMs can be a challenge in resource-constrained environments. 

Because of this, the field of model compression has become crucial, specially on LLMs \citep{zhu2023survey}. In short, model compression techniques aim to reduce either the space required to load a model, the computational cost of performing the forward pass for inference, or both. Current model compression techniques can be grouped in four types: (i) Quantization-based techniques involve reducing the precision of the parameters and/or activations \citep{liu2023llm,kim2024memory,yao2022zeroquant,park2022lut}. (ii) Pruning techniques, first introduced in \citet{lecun1989optimal}, consist on reducing the size of the model by removing parameters or components of the model. These techniques can be divided into unstructured \citep{frantar2023sparsegpt,zhang2023pruning,sun2023simple} and structured approaches \citep{ma2023llmpruner,santacroce2023matters}, where the main difference is that unstructured approaches remove individual parameters, leading to a sparse model, whereas structured approaches remove connections or hierarchical structures, therefore preserving the overall network structure. (iii) Knowledge distillation aims to transfer knowledge from a complex (teacher) model, into a simpler (student) model \citep{DBLP:journals/corr/HintonVD15,gu2024minillm,agarwal2024onpolicy,DBLP:journals/corr/abs-2212-10670,shridhar-etal-2023-distilling}. (iv) Low-rank factorization are techniques that try to approximate a parameter matrix by a product of two lower-dimensional matrices, hence reducing the number of parameters \citep{povey18_interspeech,9157223}.

Although these works on model compression have achieved great results, they generally focus on reducing the computational size/cost of the model while maintaining the general performance as much as possible. However, this leads us to the following question: What if we want to take advantage of the capabilities that an LLM has on a specific task? In other words, \emph{is it possible to extract the minimal subset from an LLM that is able to perform a specific task in a faster, standalone manner?}

Recent Mechanistic Interpretability (MI) works have shown that specific tasks are indeed carried out by a number of components that can be localized. The field of MI aims to reverse-engineer neural networks to explain their behaviors in human-understandable components \citep{elhage2021mathematical,elhage2022toy,olsson2022context}. The current workflow for most of the works is to locate a \emph{circuit} \citep{olah2020zoom} (i.e. a subset of the model) that is responsible for a specific task by performing a series of causal interventions. For example, \citet{wang2022interpretability} show that the task of Indirect Object Identification (IOI) in GPT-2 Small is performed by a circuit composed of 26 attention heads grouped in 7 different classes, according to their role in the circuit. Likewise, \citet{hanna2023does} apply different MI techniques to show how GPT-2 Small performs the greater-than operation. Similarly, \citet{pmlr-v238-garcia-carrasco24a} localize the circuit responsible for three-letter acronym prediction, as well as interpreting how the different components of the circuit work. 

There are already starting to appear automatic circuit discovery (ACD) methods \citep{conmy2023automated,bhaskar2024finding,hanna2024have} that are able to automatically identify the underlying circuit for specific tasks with good performance. However, the focus of these works has been solely on interpretability (i.e. localizing the circuit or components that are relevant for a specific task) whereas the potential for model compression has not been explored, as these methods do not enable the extraction of the discovered circuit for its independent usage. In fact, the authors of \citet{conmy2023automated} remark that their techniques generally slow the forward pass of the LLM.

In this work, we propose a novel model pruning approach that automatically extracts the subset of the model responsible for carrying out a given task on an LLM, enabling its standalone usage without any further training or fine-tuning. Given a dataset that elicits the behavior or task of interest, our approach will automatically identify the relevant components and properly prune the LLM so that we obtain the subset of components that are able to efficiently perform the task. The resulting submodel/circuit will be (i) considerably smaller in terms of the number of parameters, reducing the required storage, (ii) require a lower number of operations, thus reducing the inference time and (iii) focused on the components that perform the specific task under study, improving our ability to understand it via MI techniques, thereby increasing the model's trustworthiness. We believe that incorporating MI into model pruning approaches will enable us to fully leverage the potential of LLMs across a wide range of applications, yielding smaller, faster, and more interpretable models. In summary, our contributions are:

\begin{itemize}
    \item We propose a novel MI-based pruning method that automatically extracts the subset of an LLM that is able to properly perform a specific task of interest. Our approach does not require any extra fine-tuning and leads to models with significantly lower size and inference time.
    \item We perform an exhaustive evaluation on three different tasks whose circuits have been already manually identified on GPT-2 Small to (i) study the effect of the hyperparameters on the resulting pruned model (ii) evaluate their accuracy, size and speed and (iii) show if the pruned model contains the important components that were manually identified in previous works. 
\end{itemize}

The rest of the paper is structured as follows: Section \ref{seq:background} presents the required background. Section \ref{sec:approach} describes our approach to automatically extract task-specific circuits from LLMs. Section \ref{sec:evaluation} showcases the approach by performing an exhaustive evaluation across different tasks and models. Finally, the conclusions are presented in Section \ref{sec:conclusion}.

\section{Background}\label{seq:background}

In this section, we will briefly present the required background and techniques that will be used to efficiently extract task-specific circuits from LLMs.

\subsection{Transformer Architecture}

Practically all LLMs such as GPT \citep{radford2019language,brown2020language}, Llama \citep{touvron2023llama,touvron2023llama2} or LaMDA \citep{thoppilan2022lamda} are based on the same transformer architecture \citep{vaswani2017attention}. We will adopt and briefly present the mathematical formulation of the transformer architecture presented in \citet{elhage2021mathematical}, as it provides a better framework when it comes to our approach.

Essentially, the input to the model is a sequence of $N$ consecutive tokens which are encoded into $x_0 \in \mathbb{R}^{N \times d}$ via a learned embedding matrix $W_E \in \mathbb{R}^{V \times d}$, $V$ being the size of the vocabulary and $d$ the internal dimension of the model. Now, the vector $x_0$ is defined as the initial value of the \emph{residual stream}, where all the components of the model read from and write to. We can think of the residual stream as a main channel which components can read and modify its contents.

There are two main types of components: attention heads, and Multi-Layer Perceptrons (MLPs). The definition of components depends on the level of granularity that we want to achieve: we could break an MLP into several neurons, or merge the different attention heads into an attention layer. In our work, we decided to take individual attention heads and MLPs as components.

Following this convention, an attention layer will be composed by $n_{head}$ different heads that will independently read from the residual stream $x_i$, perform the self-attention mechanism and write the result into the residual stream. Hence, the value of the residual stream after this operation will be the sum of each contribution and the previous value of the residual stream:

\begin{equation}
    x_{i+1} = x_i + \sum_{j=0}^{n_{head}}h_j(x_i)
\end{equation}

where $h_j$ represents the output of the $j$th attention head, which performs self-attention \cite{vaswani2017attention} on $x_i$. An MLP layer $m$ will project the input to a larger space, apply a non-linearity, and project back into the residual space:

\begin{equation}
\begin{aligned}
    x_{i+2} &= x_{i+1} + m(x_{i+1}) \\ 
            &= x_{i+1} + W_2(\sigma(W_1x_{i+1}+b_1))+b_2
\end{aligned}
\end{equation}

Therefore, a sequence of an attention layer followed by an MLP will be consecutively applied until the final residual stream vector $x_L$ is unembedded via a unembedding matrix $W_U \in \mathbb{R}^{d \times V}$. Performing this unembedding operation leads to the vector $y \in \mathbb{R}^{N \times V}$ where $y_{ij}$ represents the logits of the $j$th token of the vocabulary for the prediction following the $i$th token of the sequence.

Furthermore, the computational graph of any model can be represented as a Directed Acyclic Graph (DAG), where nodes represent activations and edges represent computations between these activations. Most of the works on MI represent the models as a DAG and identify a subgraph that is responsible for the task under study, i.e. the underlying circuit. 

\subsection{Activation Patching}

Activation patching, first presented in \citet{meng2022locating}, consists on \emph{patching} (i.e. replacing) the activations of a given component with other values that disrupt its behavior. If patching the activation of a given component does not cause a significant drop in performance, it implies that such component is not relevant to the task under study, hence enabling us to locate the circuit.

There are several versions of patching that have been used across the MI literature. In our work, we focus on two of the most common, namely \emph{zero} and \emph{mean} ablation. Zero ablation, which has been suggested as a gold standard for interpretability research \citep{mcgrath2023hydra}, consists on simply setting the activations of a given component to zero, whereas mean ablation \citep{wang2022interpretability} replaces the activations with their average value across some reference distribution.

\section{Extracting Task-Specific Circuits from Large Language Models}\label{sec:approach}

As previously-mentioned, the aim of our approach is to extract a submodel from an LLM that is able to properly perform a specific task of interest. Even though there are already automated circuit discovery algorithms (ACDs) which are able to identify most of the underlying circuit for a specific task, these approaches fail to extract it so that it can be used in a standalone manner. In fact, the forward pass is typically slower than the vanilla LLM when the circuit has been identified, and naïvely pruning every component outside of the identified circuit yields disastrous results.

In this section, we present how to leverage these discovery algorithms for extracting the identified circuits enabling its independent usage. In other words, obtaining the minimal submodel (that includes the circuit) that is able to perform the specific task with a greatly reduced number of parameters and increased speed. 

The pseudocode of our approach is presented in Algorithm \ref{alg:circuit-extraction}. Essentially, given a LLM $f_\theta$ and a dataset that elicits the specific task of interest\footnote{Refer to Appendix \ref{sec:datasets} for a further discussion on the nature and curation of this dataset.} (which is split into a patching and validation datasets $D_a$ and $D_v$), our method is able to automatically obtain a pruned model $g_\theta$ that is able to perform such task. This process will be controlled by several hyperparameters, namely the threshold $\alpha$, the type of ablation used (either zero or mean ablation) and whether or not to prune MLPs.

\begin{algorithm*}
\caption{Automatic Task-Specific Circuit Extraction}\label{alg:circuit-extraction}

\KwData{Model $f_\theta$, patching dataset $D_a$, validation dataset $D_v$,evaluation threshold $\alpha$, $ablation\_scheme$, $include\_mlps$}
\KwResult{Pruned model $g_\theta$}

$g_\theta \gets f_\theta$

\For{$layer \gets [\text{num\_layers}(f_\theta), ... 0]$}{
        
    \For{$head \gets [\text{num\_heads}(f_\theta), ... 0]$}{
        
        $g'_\theta \gets g_\theta$
        
        \Comment{Temporarily ablate head}
        
        $\text{ablate\_head}(g'_\theta, layer, head, ablation\_scheme, D_a)$
        
        \If{$\text{kl\_div}(f_\theta, g'_\theta, D_v) - \text{kl\_div}(f_\theta, g_\theta, D_v) < \alpha$}
            {
            \Comment{The node is not relevant, truly prune it}

            $\text{prune\_head}(g_\theta, layer, head, ablation\_scheme, D_a)$

            }
        
    }
    
    \Comment{Optionally prune MLPs}
    
    \If{$include\_mlps = True$} {
    
        $g'_\theta \gets g_\theta$
        
        \Comment{Temporarily ablate MLP}
        
        $\text{ablate\_mlp}(g'_\theta, layer, head, ablation\_scheme, D_a)$
        
        \If{$\text{kl\_div}(f_\theta, g'_\theta, D_v) - \text{kl\_div}(f_\theta, g_\theta, D_v) < \alpha$}
            {
            \Comment{The node is not relevant, truly prune it}

            $\text{prune\_mlp}(g_\theta, layer, head, ablation\_scheme, D_a)$

            }
    }

}
   
\Return{$g_\theta$}
\end{algorithm*}

More specifically, we first initialize $g_\theta$ as the initial model $f_\theta$ and start traversing the different components of the LLM starting at the last layer. At each step, the current component is patched (using the patching dataset $D_a$), yielding a temporarily patched model $g'_\theta$. Then, we evaluate both $g'_\theta$ and $g_\theta$ by computing the KL divergence between their predictions and the original ones from $f_\theta$ on the validation dataset $D_v$. 

The difference between the computed KL divergences $\text{KL}(f_\theta \parallel g'_\theta) - \text{KL}(f_\theta \parallel g_\theta)$ quantifies the effect that patching the current component has on the performance of the specific task. In other words, if the difference is large it implies that the component had an important role on the specific task, whereas a small difference implies that the component did not have a relevant impact, thus it can be discarded. Therefore, we can specify a threshold $\alpha$ and if the difference between the KL divergences is not larger, we can permanently prune the current component and update $g_\theta$.

This operation is performed for every component, but we can choose to prune only attention heads or include MLPs in the pruning process as well. Unlike current ACD methods, our approach truly prunes the model, obtaining a considerably smaller and potentially faster submodel that is able to perform the specific task under study. This was made possible by including the following modifications that differ from current ACD methods:

\begin{itemize}
    \item If we describe the LLM as a DAG where nodes are activations and edges are operations among these activations, typical ACD methods work by patching edges instead of nodes: this enables to identify the circuit at a finer level. However, given that current accelerators excel at parallel operations, when it comes to obtaining smaller and faster submodels, it is better to prune nodes instead of edges. Even though patching nodes instead of edges provides us with a less precise circuit, ACD methods can still be applied to the resulting submodel for its interpretability.
    
    \item As previously-mentioned, current ACD methods \emph{patch} different edges, but this does not translate to actually \emph{pruning} parts of the LLM. Our approach truly prunes the components that are not relevant to the specific task, yielding a smaller and potentially faster submodel.
    
\end{itemize}

Notice that other MI works use metrics different from the KL divergence, typically derivations of the logit difference. For example, to identify the IOI circuit \citep{wang2022interpretability}, the logit difference between the indirect object (correct name) and the subject (incorrect name) is used. However, it has been shown that the KL divergence is also able to identify the circuit \citep{conmy2023automated}, with the advantage that it is a general metric that can be applied to any single-token prediction task a generally yields more robust results.

A more detailed discussion of the design choices is presented in Appendix \ref{seq:design-choices}.

\subsection{Patching vs. pruning a component}

Current implementations of ACD methods work by adding \emph{hooks} into the LLM to modify (or \emph{patch}) specific activations. This is really convenient when it comes to identifying a circuit because it enables easy and precise manipulation of the internal activations. However, ablating a component via a hook is not the same as pruning it: when patching, the component is still in the model. This implies that a patched model will have the same size as the unpatched model, and the forward pass will even be slower due to the hooks. 

\begin{figure*}
    \centering
    \includegraphics[width=\linewidth]{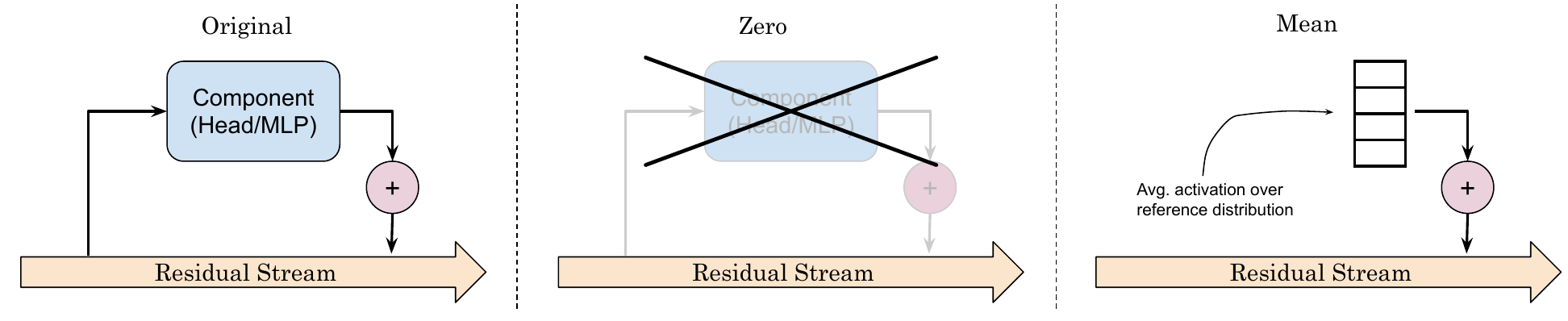}
    \caption{High-level diagram of pruning a component with zero (center) and mean (right) ablation respectively, compared with the original (left).}
    \label{fig:pruning-diagram}
\end{figure*}

On the other hand, pruning a component implies its removal from the original model, resulting in a smaller model with typically a faster forward pass. Our pruning approach depends on the type of ablation that is being used: patching with zero ablation implies replacing the activations of a component with a vector of zeros. As presented in Section \ref{seq:background}, every component in the transformer architecture reads from and writes to the residual stream in an additive manner, therefore, zero ablation is equivalent to completely removing the component. Similarly, mean ablation replaces the activations with their mean value across a reference distribution. Thus, this is equivalent to replacing the component by adding a bias term to the residual stream, greatly reducing the number of parameters and computational cost. Figure \ref{fig:pruning-diagram} shows a diagram of both types of pruning. 

This enables us to remove $4d \cdot d_{head}$ parameters for each attention head and $2d \cdot d_{mlp}$ parameters for each MLP, where $d$ is the dimension of the residual vector space, $d_{head}$ is the dimension where the head operates (typically $d = n_{head}d_{head}$, where $n_{head}$ is the number of heads per layer, and $d_{mlp}$ is the internal dimension of the MLP (typically $d_{mlp} > d$). Also, the inference time would also be reduced, as the computational costs are $\mathcal{O} (Nd_{head}d + N^2d)$ and $\mathcal{O}(Nd \cdot d_{mlp})$ for an attention head and MLP respectively, where $N$ is the size of the sequence. A more detailed mathematical formulation of each pruning method is presented in the Appendix \ref{seq:pruning}

\section{Evaluation}\label{sec:evaluation}

In order to evaluate the effectiveness of our proposal, this section will be guided by the following questions:

\begin{itemize}
    \item \textbf{RQ1:} How do the value of the threshold $\alpha$, the type of ablation and whether to prune MLPs or not affect the resulting model?
    \item \textbf{RQ2:} How much smaller is the pruned model? Is the resulting pruned model able to perform the specific task?
    \item \textbf{RQ3} Does the resulting pruned model include the circuit that performs the task under study?
    \item \textbf{RQ4:} How does our approach compare to a baseline distillation method?
\end{itemize}

Specifically, we will use our proposal to extract the underlying circuit of three different tasks whose circuit has been studied and manually identified in previous works:

\begin{itemize}
    \item \textbf{Acronym Prediction} \citep{pmlr-v238-garcia-carrasco24a}: The authors studied the task of 3-letter acronym prediction, e.g. \texttt{"The Chief Executive Officer (" $\rightarrow$ \texttt{CEO}}. Despite this being a multi-token task, we focus on the prediction of the third letter to provide a more illustrative comparison to the other tasks.  
    \item \textbf{Indirect Object Identification (IOI)} \citep{wang2022interpretability}: The authors studied the task of predicting the indirect object, e.g. \texttt{"When Mary and John went to the store, John gave a drink to " $\rightarrow$ \texttt{"Mary"}}.
    \item \textbf{Greater-than} \citep{hanna2023does}: The authors studied the ability to take in sentences such as \texttt{"The war lasted from the year 1732 to the year 17"}, and predict valid two-digit end years (years $>$ 32). 
\end{itemize}

We have chosen these tasks because their circuits have been manually identified, therefore enabling us to compare our approach to a ground truth. Similarly, we will focus on GPT-2 Small \citep{radford2019language}, as the previous works were focused on that model. 

Our method is implemented on PyTorch \citep{paszke2019pytorch} by using the TransformerLens \citep{nanda2022transformerlens} and HuggingFace transformer \citep{wolf-etal-2020-transformers} libraries. The experiments were performed on a RTX4090 GPU, on an estimated total of 72 hours of compute. 
\footnote{The code and data required to reproduce the experiments and figures, as well as the supplementary materials, can be found in \url{https://github.com/jgcarrasco/circuit-extraction}}

\subsection{RQ1: Studying the effect of the hyperparameters}

In this section, we study and provide evidence about how the different hyperparameters affect to the resulting model, namely the value of the threshold $\alpha$, whether to perform zero or mean ablation, and whether to include MLPs or not in the process. We focus on the task of acronym prediction, but the same analysis for the other two tasks can be found in Appendix \ref{seq:hypertasks}. 

Figure \ref{fig:acronyms_alpha_size} shows the impact of the threshold $\alpha$ on the size\footnote{From now on, we do not include the parameters of the embedding/unembedding matrices, as these can be thought of as lookup tables and should not be ablated.} of the resulting pruned model. The colors indicate the ablation scheme used during the extraction process (e.g. mean or zero ablation) whereas the solid/dashed lines represent whether the MLPs were not included in the pruning process (i.e. all the MLPs remain in the final submodel) or are included, respectively. 

\begin{figure}[!htbp]
    \centering
    \includegraphics[width=\linewidth]{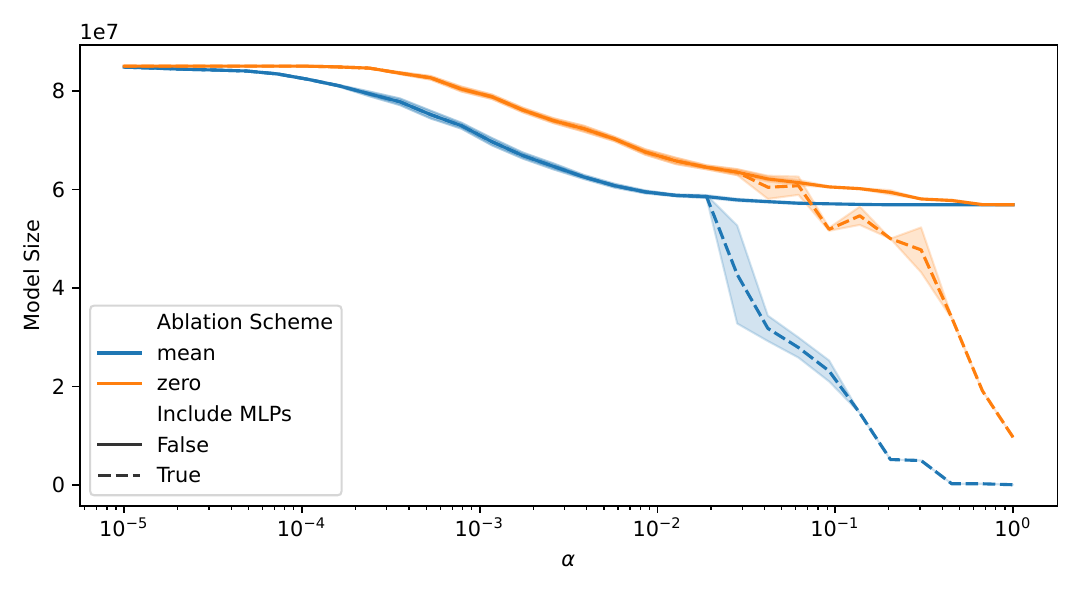}
    \caption{Impact of $\alpha$ in the resulting model size on the acronym task.}
    \label{fig:acronyms_alpha_size}
\end{figure}

The first thing to notice is that, as expected, the size of the resulting submodel decreases as $\alpha$ is set to a larger value: a larger threshold implies that for a component to be included in the final submodel it has to cause a larger increase in the KL divergence when patched. 

Another interesting fact is that, for the same value of $\alpha$, the resulting model when mean-ablating will be generally smaller than the obtained when zero-ablating. This can be explained by the fact that zero-ablation is more likely to send other components off-distribution; as every component reads from and writes into the residual stream, components that are at higher layers might be affected by the zero-ablated component, but not by the mean-ablated, as it is more likely to stay in-distribution. In other words, zero ablating a component generally causes a larger increase in KL divergence, hence more components are included into the final submodel.

When it comes to including the MLPs or not into the pruning process, it can be seen that there is no difference for lower values of $\alpha$, implying that in both cases, the attention heads are pruned first. Then, as $\alpha$ is increased, it reaches a certain point where the size of the resulting model drops abruptly, hinting that MLPs are starting to get pruned. In other words, this provides evidence that irrelevant attention heads are prioritized in the pruning process, and MLPs are starting to get pruned after all the irrelevant heads have been pruned.

\begin{figure}[!htbp]
    \centering
    \includegraphics[width=\linewidth]{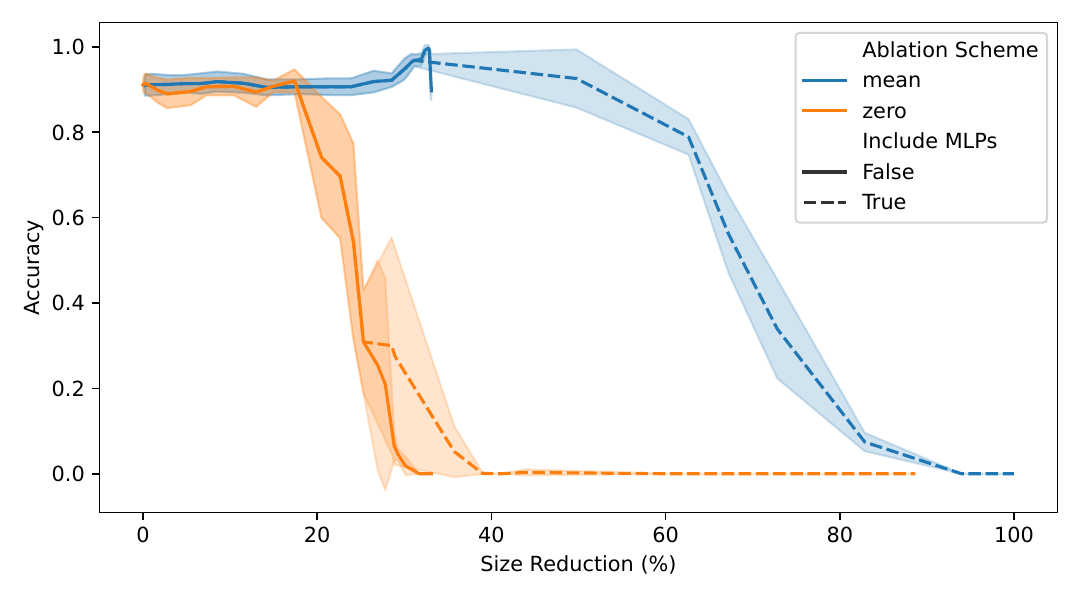}
    \caption{Accuracy vs. size of the resulting submodel on the acronym task.}
    \label{fig:acronyms_size_acc}
\end{figure}

Figure \ref{fig:acronyms_size_acc} shows the relationship between the size reduction w.r.t the unpruned model and accuracy for different hyperparameter setups. The first thing to notice is that pruning with zero ablation does not lead to any considerable size reduction without a large drop in accuracy: we can only obtain a proper submodel with up to $20\%$ size reduction, as a larger reduction hastily drops the accuracy to zero. Moreover, the results show that zero ablation yields less consistent results and are more noisy, as different executions with a fixed $\alpha$ can give models with different size and/or accuracies.

On the other hand, when the pruning is performed via mean ablation, we are able to obtain smaller submodels that are able to preserve the accuracy and even improve it. For example, in the case where MLPs are not included in the pruning process, we are able to obtain a submodel that is $33\%$ smaller, which contains just 2 attention heads and all the MLPs and has $100\%$ accuracy in the validation set. If MLPs are included in the pruning process, we are able to reduce the size by a large margin, but at the cost of some accuracy. For example, we are able to obtain a model that is $66\%$ smaller, which contains 3 attention heads and 6 MLPs and has $83.6\%$ accuracy on the validation set.

Overall, the results suggest that zero ablation might be too aggressive, requiring a large amount of components to be included in the pruned model and therefore yielding larger and slower models than when using a mean ablation scheme. On the other hand, we found out that including MLPs into the pruning process can greatly reduce the size of the resulting model, but often at the cost of performance.  

\subsection{RQ2: Performance and size comparison}\label{seq:benchmark_specific}

Table \ref{tab:comparison} shows the accuracy as well as the reduction in the number of parameters compared to the baseline for the pruned models obtained on the three different tasks. For each task, we provide two results, one where MLPs are not included in the pruning process and one where they are. The thresholds were selected according to the results of the previous section, and mean ablation is used across all runs. In order to provide more consistent results, we report the average values over five different runs with different batches of data. The size of the batches are $250$ for the acronyms and greater-than tasks and $150$ for IOI. A Table with extra information about the resulting size and inference time can be found in Appendix \ref{seq:benchmark}.

\begin{table*}[!htbp]
\centering
\begin{tabular}{lllllllll}
\toprule
Task & $\alpha$ & MLP & $acc$ (\%) & $\Delta param$ (\%) & TPR (\%) & FPR (\%) \\
\midrule
Acronyms & $8.86 \cdot 10^{-2}$ & $False$ & $99.92 \pm 0.17 $ & $32.88 \pm 0.00$ & $20.00 \pm 6.85$ & $0.30 \pm 0.40$\\
& $3.50 \cdot 10^{-2}$ & $True$ & $78.64 \pm 5.16$ & $63.70 \pm 3.10$ & $40.00 \pm 5.59$ & $0.59 \pm 0.33$ \\ \midrule
IOI & $8.53 \cdot 10^{-3}$ & $False$ & $100.00 \pm 0.00 $& $28.66 \pm 0.50$ & $57.40 \pm 1.94$ & $5.95 \pm 0.69$ \\
& $1.88 \cdot 10^{-2}$ & $True$ & $96.53 \pm 1.52$ & $72.31 \pm 2.98$ & $35.65 \pm 1.94$ & $0.99 \pm 0.37$ \\ \midrule
Greater-than & $8.53 \cdot 10^{-2}$ & $False$ & $100.00 \pm 0.00 $& $32.65 \pm 0.00$ & $37.50 \pm 0.00$ & $0.00 \pm 0.00$\\
& $8.53 \cdot 10^{-2}$ & $True$ & $99.84 \pm 0.36$ & $82.77 \pm 0.13$ & $25.00 \pm 0.00$ & $0.00 \pm 0.00$\\ \bottomrule
\end{tabular}
\caption{Evaluation of the pruned models obtained on each of the tasks for different values of $\alpha$, as well as their recovery rates when compared with the manually identified circuits. The reported results are averaged across five repetitions.}
\label{tab:comparison}
\end{table*}

The results show that our approach yields models that are able to perform the specific tasks while preserving or even improving the performance of the original LLM (see Appendix \ref{seq:benchmark} for extended results) while having a drastically reduced size and inference time. In general, including MLPs in the pruning process gives even better size and time reductions with negligible drops in performance. However, it seems that some tasks might be more affected than others, as we found that submodels with a reduction size of more than $60\%$ on the acronyms task gave a considerably lower performance of $78.64\%$ on average.

\subsection{RQ3: Comparison to manually identified circuits}

One important aspect to take into account is to check whether the pruned submodel contains the circuit that performed the specific task in the original LLM. Specifically, as shown on Table \ref{tab:comparison}, we compare the attention heads present in the pruned model with the heads that were manually identified in previous MI works and computed the True Positive Rate (TPR) and False Positive Rate (FPR).  

Interestingly, the TPRs are not very high, implying that there are attention heads what were manually identified as important that are not present in the pruned model. This is most likely caused by the following facts. First, there are manually identified heads that contribute very small amounts to the performance, hence they can often be discarded. For example, the authors of \citet{pmlr-v238-garcia-carrasco24a} show that almost all the performance on the acronyms task can be recovered with just 2/3 heads, where the circuit is composed by 8 heads. Second, previous works might also use other kinds of ablation (e.g. resampling ablation), whereas we use mean ablation. This implies that heads that were deemed as important via other ablation schemes can be approximated by their mean output across a reference distribution (e.g. mean ablation) and therefore discarded.  There are also other factors that have an effect such as the metric choice, the order of ablation, or the human factor. However, the FPRs are extremely low, suggesting that the pruned models contain the most important heads that were also discovered manually in previous works, which is actually what matters. Appendix \ref{sec:roc} provides a further discussion on this, including the ROC curves for each task.

\subsection{RQ4: Comparison to a baseline}

Even though our approach is built upon a completely different motivation (i.e. extracting interpretable task-specific circuits from an already pretrained model), it is interesting to compare our results with a baseline involving task-specific model compression. Specifically, we will focus on the technique of model distillation \citep{ba2014deep}, which enables transferring information from a teacher, larger network, to a student, smaller network. 

We adopt a similar setup to \citet{tang2019distilling}. Essentially, they train the student model to output the same same logits than the teacher model across a given training dataset of samples:

\begin{equation}
    \mathcal{L}_{distill} = || \mathbf{z}^{(T)} - \mathbf{z}^{(S)} ||^2_2
\end{equation}

where $\mathbf{z}^{(T)}$ and $\mathbf{z}^{(S)}$ are the teacher and student logits, respectively. To provide a fair comparison, the teacher network  will be the same GPT-2 Small pretrained model as on the previous experiments. Regarding the student network, it will have the same transformer architecture, but with a smaller number of layers. The model is trained by minimizing $\mathcal{L}_{distill}$ for a total of $20000$ epochs with the Adam optimizer \citep{kingma2014adam} and a learning rate of $10^{-3}$.

\begin{table}[!htb]
\centering
\begin{tabular}{lllll}
         & & \textbf{Distillation} &   \textbf{Ours}\\ \toprule
    Task & $\Delta param$ (\%) & $acc$ (\%) &  $acc$ (\%)\\
    \midrule 
    A & $32.88 \pm 0.00$ & $12.53 \pm 6.27 $ &  $\mathbf{99.92 \pm 0.17}$\\
      & $63.70 \pm 3.10$ & $10.66 \pm 2.89$ &  $\mathbf{78.64 \pm 5.16}$ \\ \midrule
    IOI & $28.66 \pm 0.50$ & $2.40 \pm 0.33 $ &  $\mathbf{100.00 \pm 0.00}$\\
      & $72.31 \pm 2.98$ & $ 2.13 \pm 0.65 $ & $\mathbf{96.53 \pm 1.52}$ \\ \midrule
      GT & $32.65 \pm 0.00$ & $77.60 \pm 32.11$ &  $\mathbf{100.00 \pm 0.00}$\\
      & $82.77 \pm 0.13$ & $93.20 \pm 2.84$ &  $\mathbf{99.84 \pm 0.36}$ \\ \bottomrule
\end{tabular}
\caption{Comparison of the performance and size of the pruned model obtained with our approach and the performance of a student network with a similar size obtained by knowledge distillation from GPT-2 Small.}
\label{tab:baseline}
\end{table}

Table \ref{tab:baseline} presents the results of our approach versus the knowledge distillation approach for each of the three tasks of study. Specifically, for each of the previous pruned models obtained with our approach, we initialize the student model as an $N$-layer transformer such that $N$ is the largest number such that the size of the student model is less or equal than the size of our resulting pruning model. Then, we start the knowledge distillation procedure and report the largest accuracy obtained in the validation dataset. 

The results clearly show that the smaller models obtained by distillation are unable to perform the specific task, with the exception of the greater-than task. The main factor behind this result is that we have used a small amount of data samples, which are not enough for training a model from scratch but sufficient to properly extract the subcircuit via our method. 

\section{Conclusion}\label{sec:conclusion}

Recent works in MI have made impressive advances in circuit identification to localize task specific behaviors on LLMs. However, current methods ae not able to extract such circuits to enable its standalone usage, thereby missing the benefits of reduced size and inference time.

In this work, we proposed a novel circuit extraction method that, given a datset that elicits a specific task of interest, automatically prunes the LLM to obtain a minimal subset capable of performing the task without additional training or fine-tuning. We extensively evaluated our method on three tasks, whose underlying circuits have already been manually identified, demonstrating that the pruned models obtained with our method are (i) able to properly perform the specific task, often better than the original LLM (ii) considerably smaller than the original LLM, with up to $82.77\%$ reductions in size and (iii) align with most results from previous literature on circuit identification, as they do not include components that were deemed as irrelevant on previous works. We also compared our approach with a distillation method and showed that the models obtained via distillation were not able to properly perform the tasks due to requiring larger amounts of training data, whereas our approach is able to extract circuits with small amounts of data.

To the best of our knowledge, this is the first work that tries to automatically extract task-specific circuits from LLMs for its standalone usage as well as the first work that proposes a MI-based pruning approach. In an era where the size of the models is getting exponentially larger, the ability to distill task-specific subsets capable of performing with significantly reduced computational overhead becomes increasingly vital. Moreover, deploying entire LLMs not only incurs unnecessary computational costs but also retains irrelevant components, exacerbating their black-box nature and potentially introducing security vulnerabilities. Our work tries to pave the way into leveraging the power of LLMs by extracting task-specific circuits that are able to perform faster, as well as being more interpretable. We believe this to be a crucial step towards more efficient, trustworthy and efficient AI systems.
  
Our work is limited to just one model and three fixed sequence length, single-token prediction tasks, in order to evaluate it on the current MI literature. However, these results show that extracting task-specific circuits from LLMs is possible. Future work includes extending the evaluation to more complex tasks such as multi-token prediction, generative tasks or sentiment analysis as well as use larger, production-grade LLMs. 

\section*{Acknowledgements}

This work has been supported by the AETHER-UA project (PID2020-112540RB-C43) funded by Spanish Ministry of Science and Innovation. Jorge García-Carrasco holds a predoctoral contract (CIACIF / 2021 / 454) granted by the Conselleria de Innovación, Universidades, Ciencia y Sociedad Digital (Generalitat Valenciana). This work is part of the TSI-100927-2023-6 Project, funded by the Recovery, Transformation and Resilience Plan from the European Union Next Generation through the Ministry for Digital Transformation and the Civil Service.

\bibliography{aaai25}

\clearpage
\appendix
\section*{Appendix}
\section{Datasets}\label{sec:datasets}

The current workflow on circuit identification is to first build a synthetic dataset that elicits the behavior or specific task under study. Such dataset is not used for training nor fine-tuning, but to identify the relevant components of the circuit via a series of causal interventions, or activation patching experiments. The dataset curation process is specific to the task under study, but previous works follow these general guidelines or suggestions:

\begin{itemize}
    \item The different prompts are built according to a predefined template so that, when tokenized, the important tokens (e.g. the first letter of a noun in the acronym prediction task) share the same position across all samples. The reason behind is that most ablation schemes work by replacing the activations from other samples, implying that (i) we are unable to use sequences with different lengths and (ii) it is recommended that they share the same template, so that the semantic meaning is distributed similarly across samples.
    \item Words or letters that are thought to be relevant for the specific task are tokenized individually in a single token. In other words, in the hypothetical task of predicting names from a sentence, it is recommended to build sentences with names that are tokenized as a single token (e.g. \texttt{"|Mary|"}).
\end{itemize}

However, it is important to remark that these are suggestions to ease the process of identifying the circuit and understanding the different components. In fact, previous works have built datasets following different templates and found the same underlying circuit.  

In our experiments, we will use the same datasets that were built for the purpose of manual identification of circuits responsible for different tasks on GPT-2 Small. Below we present a description of each dataset:   

\begin{itemize}
    \item \textbf{Acronym Prediction: } As the task under study is the prediction of 3-letter acronyms, the dataset is composed by prompts such as \texttt{"The Chief Executive Officer (CEO"}, where the task would be to predict the acronym. More especifically, the prompts share the same underlying template: \texttt{\texttt{"|The|C1|T1|C2|T2|C3|T3| (|A1|A2|A3|"}}, where \texttt{Ci} is the token encoding the capital letter of the $i$th word (together with its preceding space), \texttt{Ti} is the remainder of the word, and \texttt{Ai} is the $i$th letter of the acronym. Despite this being a multi-token prediction task, we limit our analyisis to single-token by focusing on the prediction of the last letter of the acronym (i.e. \texttt{A3}) in order to provide a better comparison with the other tasks, which are single-token prediction tasks.
    \item \textbf{Indirect Object Identification (IOI): } The dataset is composed by a total of 15 templates. However, we focus on just one template for the sake of giving a more illustrative comparison and evaluation of our proposal. Specifically, we focus on the following template: \texttt{"Then, [B] and [A] went to the [PLACE]. [B] gave a [OBJECT] to [A]"}, where \texttt{[A]} and \texttt{[B]} are first names in English. The first names, places and objects are selected so that they are encoded as a single token each to ensure proper alignment across sequences.
    \item \textbf{Greater-Than: } The dataset is composed by sentences that follow the template: \texttt{"The [noun] lasted from the year XXYY to XX"}, where \texttt{[noun]} is drawn from a hardcoded pool of nouns (e.g.  abduction, accord, affair, etc.) and \texttt{XXYY} is a year, where \texttt{XX} represents the century, which is drawn from $\{11,...,17\}$, and \texttt{YY} represents the rest of the year, which is drawn from $\{02,...,98\}$. The reason behind this design choice is that there are years, such as 1700, which are naturally tokenized by GPT-2 Small tokenizer as a single token. Hence, both the century and start year have to be carefully sampled so that the resulting years are tokenized as \texttt{|XX|YY|}.
\end{itemize}

A more thorough explanation of the dataset building process can be found in their respective works \citet{pmlr-v238-garcia-carrasco24a,wang2022interpretability,hanna2023does}.

\section{Design Choices}\label{seq:design-choices}

In this section, we provide a further discussion on the design choices of Algorithm \ref{alg:circuit-extraction} as well as additional information.

\subsection{Nodes vs. Edges}

Current ACD methods and most MI works focused on circuit identification represent LLMs as a Directed Acyclic Graph (DAG), where nodes represent activations (e.g. the output of an attention head or an MLP) and edges represent computations (e.g. computing the attention patterns). The main idea of current ACD methods is to patch unimportant edges: if patching an edge does not drop the performance over a specified threshold, it implies that it is not important for the specific task and it is discarded. 

Patching edges instead of nodes usually results in a more fine-grained and precise circuit: by removing all unimportant edges, one is able to clearly see how the remaining nodes are connected (e.g. the output $8th$ attention head of the $9th$ layer is connected to the query input of the $5$th attention head of the $11th$ layer). While this is a desirable aspect when solely focusing on interpretability, we encounter two main challenges when we focus on a model pruning standpoint (i) current MI works do not actually remove the edges/operations, they are patched via hook functions that actually slow the forward pass even more and (ii) it is difficult to efficiently translate these removed edges and truly prune them, as current transformers are implemented via large parallel operations in the shape of matrix multiplications.

Because of this, we focus on nodes instead of edges: the remaining circuit will be less precise (i.e. it will contain many more edges) but we will be able to efficiently prune it, obtaining a faster and reduced submodel that is able to perform the task under study. However, this should not be seen as an inconvenience: ACD algorithms can also be applied to our obtained submodel to obtain a more fine-grained circuit for interpretability purposes and keep our submodel to perform faster inference.

\subsection{Zero vs. Mean ablation schemes} 

Zero ablation implies replacing the output of a component (i.e. an attention head or MLP) by a tensor of zeros, whereas mean ablation implies replacing the output by the mean tensor obtained on a reference distribution (i.e. the patching dataset). 

As mentioned in the paper, zero ablation is regarded as a more ``aggressive'' ablation with regards to mean ablation in the MI literature. The intuition behind this is that, due to every component of a transformer model reading from and writing to a common residual stream, zero-ablating a irrelevant component could send components from upper layers off-distribution which might be relevant for the task under study. On the other hand, mean-ablation will cause the component to write common information from the reference distribution, hence being considerably less likely to send other components off distribution.

Even though our initial thoughts before the experiments tended toward mean ablation, we also decided to perform experiments with zero ablation and found the expected results. In summary, zero ablation is too aggressive and therefore requires a larger number of components to be included in the final submodel in order to maintain a good performance.

\subsection{KL Divergence vs. other metrics}

Previous works on manual circuit identification used metrics different from the KL divergence, mostly variations of the logit difference. However, the authors of \citet{conmy2023automated} extensively evaluated ACD with different metrics and tasks and found out that the KL divergence was the most effective, yielding more consistent results. They also show that some metrics might peform better on the discovery of specific task, which is something that should be a focus of further research. Nevertheless, as the aim of our work is to perform automatic circuit extraction, we decided to only use the KL divergence to provide a more illustrative evaluation and leave the metric selection aspect for further works.  

Barrier
\section{Truly Pruning Attention Heads}\label{seq:pruning}

To understand how to truly prune attention heads, we first have to present how attention layers are generally implemented. 

First, the input to the attention layer is a residual stream tensor $x \in \mathbb{R}^{B \times N \times d}$, where $B$ is the batch size, $N$ is the sequence length and $d$ is the model dimension. As described in \citet{vaswani2017attention}, this vector $x$ is mapped into three different vectors termed the query, key and value vectors via a linear mapping. Tipically, this projection operation is implemented in parallel for every attention head in the layer:

\begin{equation}
    \text{Concat}(q, k, v) = x W_{proj} = x \text{Concat}(W_Q, W_K, W_V)
\end{equation}

where $W_Q, W_K, W_V \in \mathbb{R}^{d \times d}$. This operation yields the $q,k,v \in \mathbb{R}^{B \times N \times d}$ vectors which are then split into $n\_{head}$ Q, K, V vectors associated to the $i$th attention head $q_i,k_i,v_i \in \mathbb{R}^{B \times N \times d_{head}}$, where $d = n_{head} d_{head}$. Then, these vectors are used to compute the output of each attention head $h_i \in \mathbb{R}^{B \times N \times d_{head}}$. 

Finally, the results of every attention head are stacked and projected together as follows:

\begin{equation}
    \text{MultiHeadAttention}(x) = \text{Concat}(h_1,h_2,...,h_n)W^O
\end{equation}

where $W^O \in \mathbb{R}^{n_{head} \cdot d_{head} \times d}$ is a projection matrix. However, this is equivalent to projecting each head independently and then summing the individual contributions:

\begin{equation}
    \text{MultiHeadAttention}(x) = \sum_{i=0}^n h_i W^O_i
\end{equation}

where $W^O_i \in \mathbb{R}^{d_{head} \times d}$. Therefore, the matrices $W_Q, W_K, W_V$ can be represented as a stack of matrices $W_Q = [W_Q^1, W_Q^2, ...W_Q^{n_{head}}]^T$, where $W_Q^{i} \in \mathbb{R}^{d \times d_{head}}$ would be the projection matrix to obtain the Q vector for the $i$th attention head $q_i$, and similarly for the K and V vectors. Moreover, we also have that $W_O = [W_O^1, W_O^2, ...W_O^{n_{head}}]$. 

Hence, zero ablating the $i$th head directly translates to removing the $W_Q^i, W_K^i, W_V^i$ and $W_O^i$ matrices. On the other hand, mean ablating implies removing the previous matrices and adding the mean output vector across a reference distribution to the output of the pruned layer. In contrast to current ablating implementations used on MI works which are based on hooks, our pruning process truly removes the component, yielding the benefit of size and inference time reduction.

\section{Study of the effect of hyperparameters on other tasks}\label{seq:hypertasks}

\begin{figure}
    \centering
    \includegraphics[width=\linewidth]{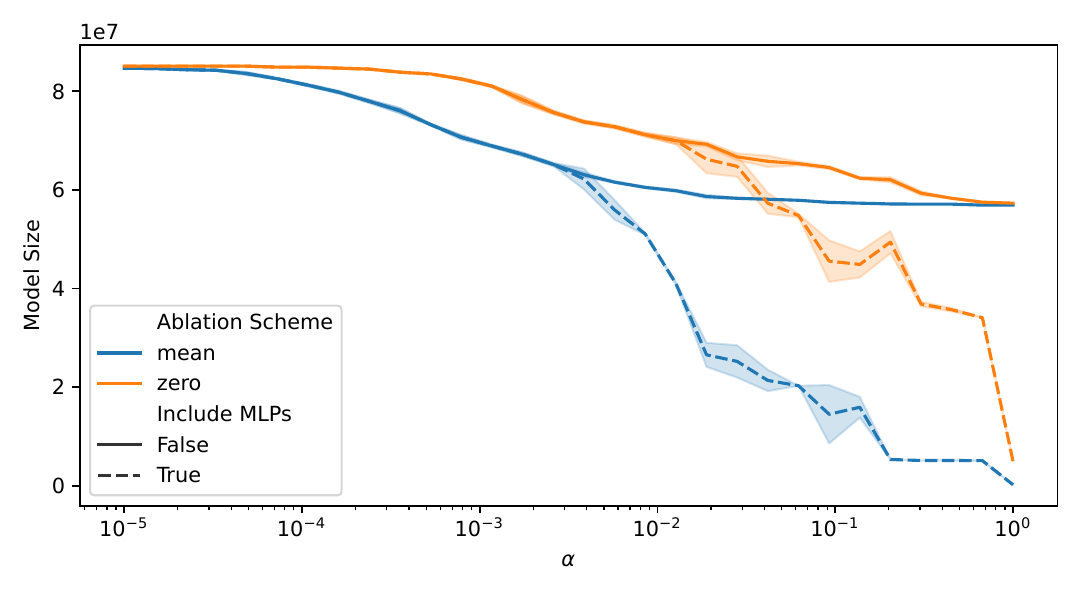}
    \caption{Impact of $\alpha$ in the resulting model size on the IOI task.}
    \label{fig:ioi_alpha_size}
\end{figure}

\begin{figure}
    \centering
    \includegraphics[width=\linewidth]{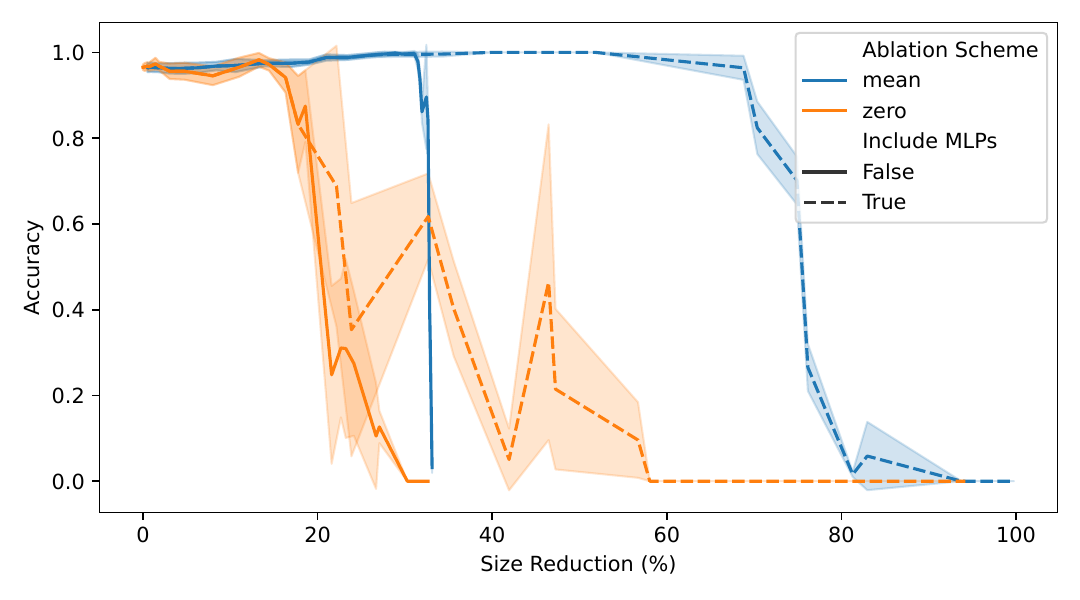}
    \caption{Accuracy vs. size of the resulting submodel on the IOI task.}
    \label{fig:ioi_size_acc}
\end{figure}

\begin{figure}
    \centering
    \includegraphics[width=\linewidth]{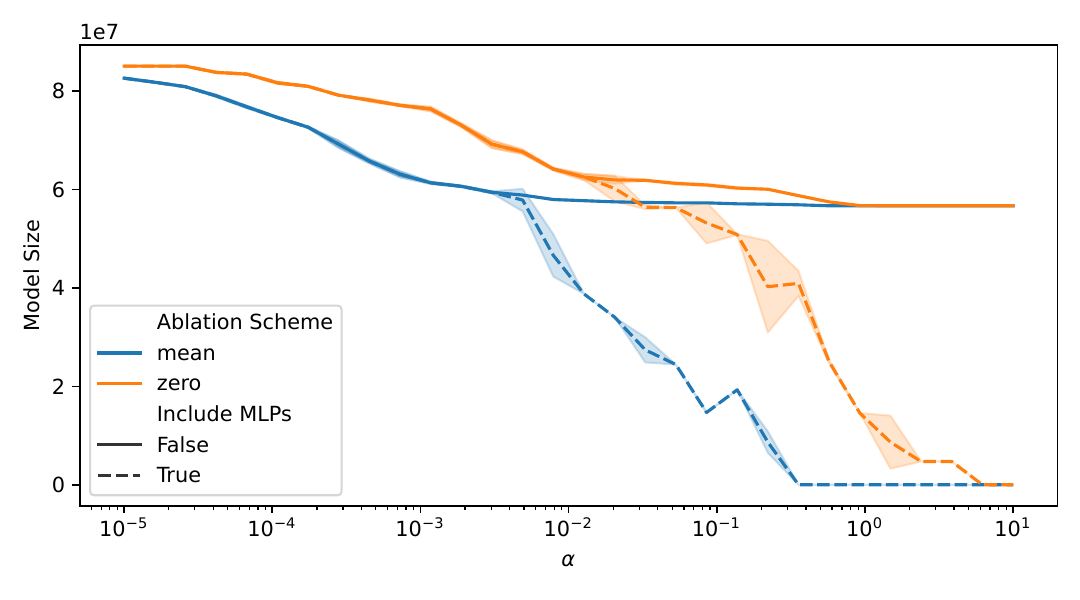}
    \caption{Impact of $\alpha$ in the resulting model size on the greater-than task.}
    \label{fig:greater_alpha_size}
\end{figure}

\begin{figure}
    \centering
    \includegraphics[width=\linewidth]{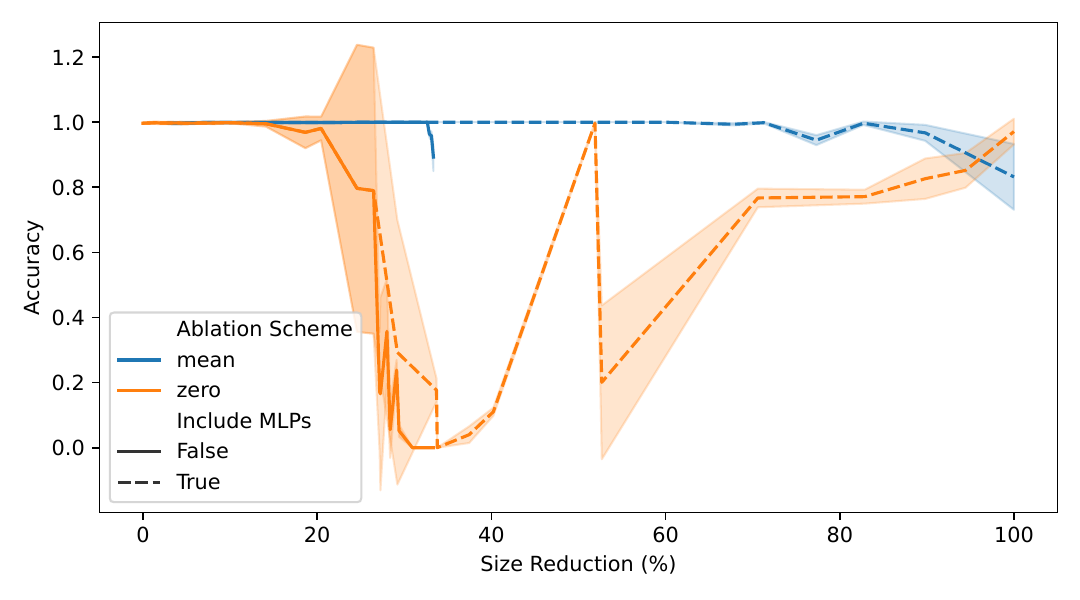}
    \caption{Accuracy vs. size of the resulting submodel on the greater-than task.}
    \label{fig:greater_size_acc}
\end{figure}

Figures \ref{fig:ioi_alpha_size} and \ref{fig:ioi_size_acc} show the impact of $\alpha$ on the size and the accuracy vs. the size of the resulting submodel on the IOI task. Similarly, Figures \ref{fig:greater_alpha_size} and \ref{fig:greater_size_acc} show the same results on the greater-than task. Overall, we can draw the same conclusions that we obtained on the acronyms task: First, for a fixed $\alpha$, mean ablation yields smaller and therefore faster models than zero ablation. Second, the results when using zero ablation are considerably less consistent, as shown by the high error bars, specially on the accuracy vs. size plots. Finally, there is no difference between including MLPs or not at lower threshold levels $\alpha$, implying that the pruning of irrelevant attention heads are prioritized over irrelevant MLPs.

Another interesting result can be found in Figure \ref{fig:greater_size_acc}. Specifically, it can be seen that when performing zero ablation and including MLPs in the greater-than task, the performance abruptly increases to 100\% when the resulting pruned model is around $50\%$ of the original size, as well as approaching the 100\% reduction size. This is most likely due to the fact that the KL divergence is not correlated with how the accuracy is computed: if the model always outputs the starting year 99, the accuracy will be 100\% in every case, but these predictions would have a large KL divergence with respect to the original distribution. While this is an interesting phenomena to study, it is out of scope of this paper and leave it out for further research.
 
Barrier
\section{Extra Benchmark Results}\label{seq:benchmark}

Table \ref{tab:comparison_extended} shows the extended results from the evaluation presented in Section \ref{seq:benchmark_specific}. 

\begin{table}[!h]
\footnotesize
    \centering
    \caption{Extended results of the evaluation of the pruned models obtained on each of the tasks for different values of $\alpha$. The process is repeated across five different batches and the results are averaged.}
    \label{tab:comparison_extended}
    \rotatebox{90}{ 
    \begin{tabular}{lllllllll}
    \toprule
        Task & $\alpha$ & MLP & $acc$ (\%) & $\Delta acc$ (\%) & \# param. $(10^7)$ & $\Delta param$ (\%) & t (ms) & $\Delta t$ (\%)\\
        \midrule 
        Acronyms & $8.86 \cdot 10^{-2}$ & $False$ & $99.92 \pm 0.17 $ & $9.28 \pm 1.53$ & $5.70 \pm 0.00$ & $32.88 \pm 0.00$ & $1.56 \pm 0.02$ & $83.56 \pm 0.53$\\
          & $3.50 \cdot 10^{-2}$ & $True$ & $78.64 \pm 5.16$ & $-12.00 \pm 5.5$ & $3.09 \pm 0.26$ & $63.70 \pm 3.10$ & $1.32 \pm 0.12$ & $86.20 \pm 1.02$\\ \midrule
        IOI & $8.53 \cdot 10^{-3}$ & $False$ & $100.00 \pm 0.00 $ & $2.93 \pm 1.60$ & $6.07 \pm 0.04$ & $28.66 \pm 0.50$ & $3.10 \pm 0.65$ & $57.58 \pm 8.76$\\
          & $1.88 \cdot 10^{-2}$ & $True$ & $96.53 \pm 1.52$ & $-1.73 \pm 2.14$ & $2.35 \pm 0.25$ & $72.31 \pm 2.98$ & $1.26 \pm 0.05$ & $82.78 \pm 0.72$\\ \midrule
          Greater than & $8.53 \cdot 10^{-2}$ & $False$ & $100.00 \pm 0.00 $ & $0.00 \pm 0.00$ & $5.73 \pm 0.00$ & $32.65 \pm 0.00$ & $1.79 \pm 0.01$ & $77.78 \pm 0.15$\\
          & $8.53 \cdot 10^{-2}$ & $True$ & $99.84 \pm 0.36$ & $-0.08 \pm 0.04$ & $1.47 \pm 0.01$ & $82.77 \pm 0.13$ & $0.94 \pm 0.08$ & $88.33 \pm 1.00$\\ \bottomrule 
          
    \end{tabular}}
\end{table}

Barrier
\section{Comparison to Manual Circuit Identification}\label{sec:roc}

\begin{figure}[!htbp]
    \centering
    \includegraphics[width=\linewidth]{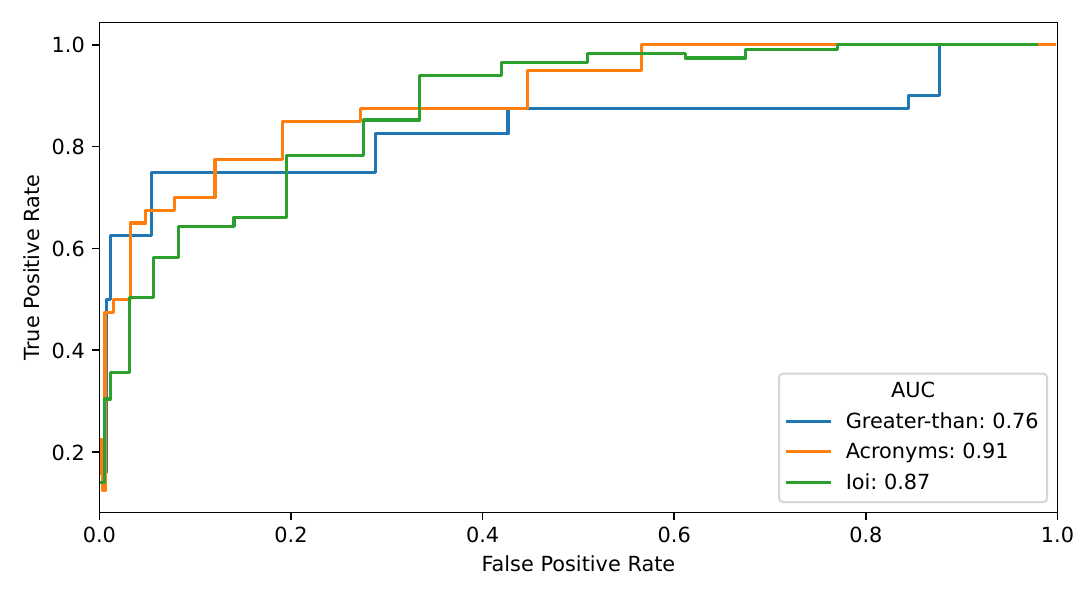}
    \caption{ROC curves of the identified attention heads for the three tasks of study. Mean-ablation is used.}
    \label{fig:pareto-mean}
\end{figure}

Figure \ref{fig:pareto-mean} shows the True Positive Rates (TPRs) and False Positive Rates (FPRs) for the discovered attention heads obtained by varying the threshold and applying the mean-ablation scheme. In general, we can see that our method is able to automatically recover most of the components. The results with lower AUC are obtained in the greater-than task. This is most likely due to the nature of the task under study. Differently from the other two tasks that we have analyzed, in the greater-than task there are more than one possible correct answer (i.e. any start year greater than the one in the sentence). The authors use a task-specific metric that takes this into account wereas we stick to the general KL divergence. However, our method is able to include the most relevant components and the resulting submodel is able to maintain and even improve the accuracy with a considerably smaller size.

\begin{figure}[!htbp]
    \centering
    \includegraphics[width=\linewidth]{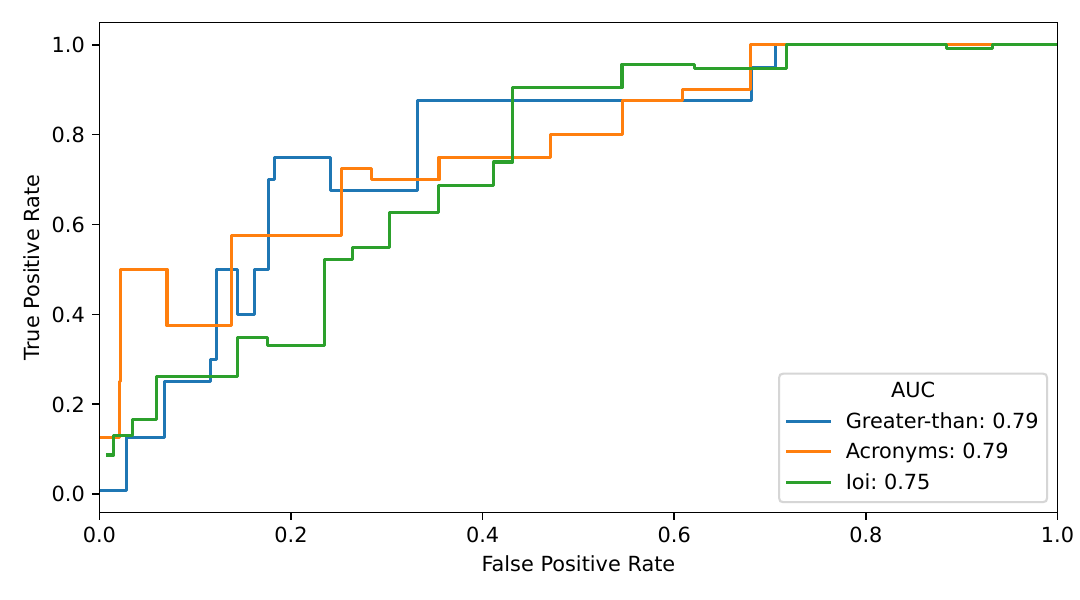}
    \caption{ROC curves of the identified attention heads for the three tasks of study. Zero-ablation is used.}
    \label{fig:pareto-zero}
\end{figure}

Figure \ref{fig:pareto-zero} shows the results by using a zero-ablation scheme, instead of replacing by the mean activations. As expected, we can see that the AUCs are generally lower. This is due to the fact that the main objective in previous MI works was to identify the principal components that are responsible for a specific task. However, these components rely on other upstream (or lower-layer) components that might not be relevant for the task, but belong to other ``sub-circuits'' that support the main circuit. For example, in the acronym task, the authors found that some positional information used by the circuit was propagated by other upstream attention heads that did not belong to the circuit. As zero-ablation is equivalent to completely removing a component (i.e. it is more ``aggressive''), the components that might not be directly in the circuit but in a supportive way are included.

\end{document}